\documentclass[letterpaper, 10 pt, conference]{ieeeconf}  

\IEEEoverridecommandlockouts                              

\usepackage{times}
	\usepackage{graphicx,times}
	\usepackage{subfigure}
\usepackage{amsmath}
\usepackage{amssymb}
\usepackage{mathrsfs}
\usepackage{multirow}
\usepackage{booktabs}
\usepackage{makecell}

\usepackage{algorithm, algorithmicx, algpseudocode}

\usepackage{array}
\usepackage{url}
\usepackage{mathptmx} 
\newcolumntype{P}[1]{>{\centering\arraybackslash}p{#1}}
\newcolumntype{M}[1]{>{\centering\arraybackslash}m{#1}}

\title{\LARGE \bf
SD-Net: Symmetric-Aware Keypoint Prediction and Domain Adaptation for 6D Pose Estimation In Bin-picking Scenarios
}

\author{Ding-Tao Huang$^{1,2,\dag}$, En-Te Lin$^{1,\dag}$, Lipeng Chen$^{2,\dag}$, Li-Fu Liu$^{1}$, Long Zeng$^{1*}$ %
\thanks{\dag \ Equal contribution. }
\thanks{$^{1}$ International Graduate School at Shenzhen, Tsinghua University, Shenzhen, China.  This work was partially done when Dingtao Huang conducted an internship at Tencent. {\tt\small \{hdt22, linet22, llf23\}@mails.tsinghua.edu.cn; zenglong@sz.tsinghua.edu.cn.} }
\thanks{$^{2}$ Tencent Robotics X, Shenzhen, China. {\tt \small lipengchen@tencent.com%
}}
\thanks{$^{*}$  Corresponding author.}
}
		
\begin{document}

\maketitle

\begin{abstract}

Despite the success in 6D pose estimation in bin-picking scenarios, existing methods still struggle to produce accurate prediction results for symmetry objects and real world scenarios. The primary bottlenecks include 1) the ambiguity keypoints caused by object symmetries; 2) the domain gap between real and synthetic data. To circumvent these problem, we propose a new 6D pose estimation network with symmetric-aware keypoint prediction and self-training domain adaptation (SD-Net). SD-Net builds on pointwise keypoint regression and deep hough voting to perform reliable detection keypoint under clutter and occlusion. Specifically, at the keypoint prediction stage, we designe a robust 3D keypoints selection strategy considering the symmetry class of objects and equivalent keypoints, which facilitate locating 3D keypoints even in highly occluded scenes. Additionally, we build an effective filtering algorithm on predicted keypoint to dynamically eliminate multiple ambiguity and outlier keypoint candidates. At the domain adaptation stage, we propose the self-training framework using a student-teacher training scheme. To carefully distinguish reliable predictions, we harnesses a tailored heuristics for 3D geometry pseudo labelling based on semi-chamfer distance. On public Sil\'eane dataset, SD-Net achieves state-of-the-art results, obtaining an average precision of 96\%. Testing learning and generalization abilities on public Parametric datasets, SD-Net is 8\% higher than the state-of-the-art method. The code is available at \url{https://github.com/dingthuang/SD-Net}.

\end{abstract}

\section{Introduction} \label{introduction}

The estimation of 6D object pose is an essential prerequisite for robotic tasks such as grasping and manipulation, especially in bin-picking scenarios \cite{opnet1}. Recent studies that employ learning-based techniques have shown promising results for this particular task \cite{opnet2}. These methods primarily fall into two categories: holistic methods \cite{opnet1,p1} and keypoint-based methods \cite{pvnet,pvn3d}. Keypoint-based methods employ intermediate variables to predict the 6D object pose, effectively circumventing the nonlinear rotation space and providing a promising direction for the exploration of 6D Object Pose Estimation \cite{duan1}.
\begin{figure}[t]
	\centering
		\includegraphics[width=1.0\columnwidth]{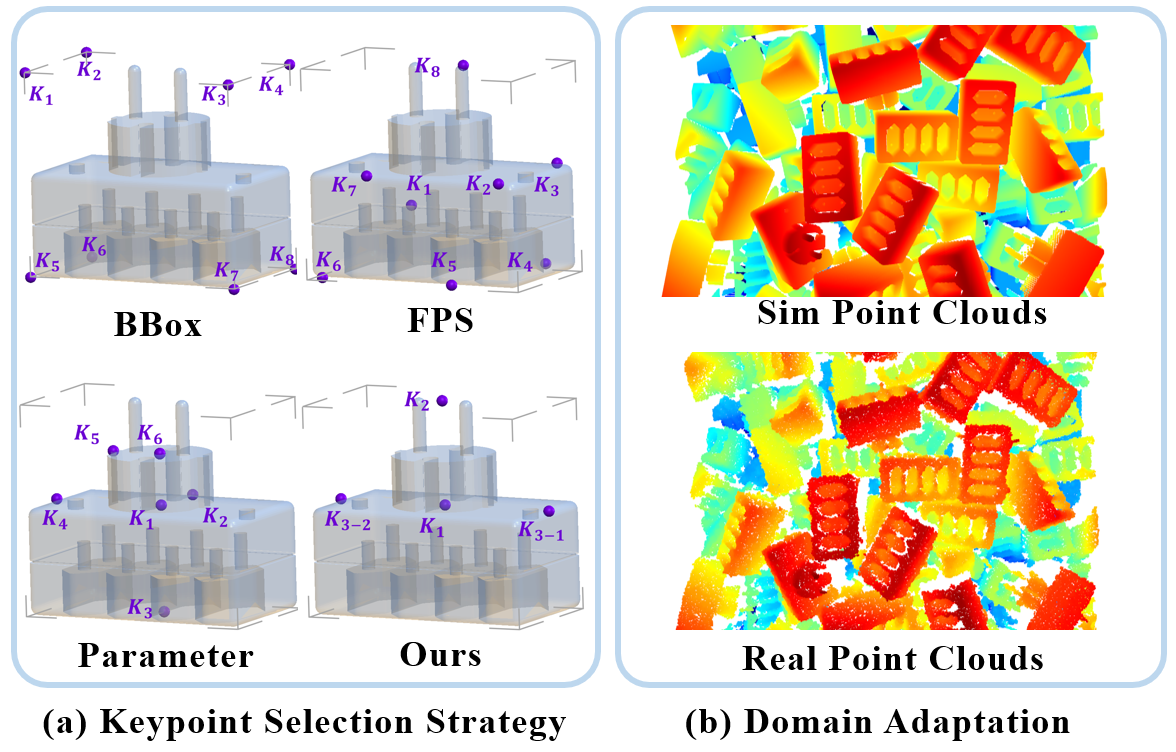}
	\caption{This paper addresses two major problems facing 6D pose estimation. (a) Keypoints sampled by BBox, FPS, Parameter and ours strategy. BBox, FPS and Parameter keypoints ignore object geometric symmetry characteristics, while ours strategy adaptively select keypoints based on object symmetry class. (b) Methods typically exhibit inferior performance when applied to real world point clouds, due to the persistent domain gap between real and synthetic data point clouds.}
	\label{fig:fig1}
\end{figure}

Nonetheless, two major challenges persist in hindering the estimation performance of keypoint-based pose estimation approaches, as illustrated in Fig. \ref{fig:fig1}. The first challenge lies in the absence of a robust strategy for selecting keypoints during the keypoint prediction stage. This deficiency is particularly evident when dealing with objects where keypoints on symmetric components are hardly distinguishable. Many previous works \cite{Bb8,YOLO-6D} have chosen keypoints from a subset of the object's bounding box (BBox) corners which away from the surface of objects. Some other works \cite{pvn3d, pvnet} employs the farthest point sampling (FPS) algorithm to sample keypoints on the object's surface according to their relative proximity. ParametricNet \cite{parametricnet} utilizes a keypoint selection strategy based on object parameters. Despite promising results, these methods fail to consider object geometric symmetry characteristics. The symmetry of an object can result in multiple points on the surface that have similar geometric features to the selected keypoints. These ambiguity points mislead the network in predicting keypoints.

Secondly, annotating 6D object poses in the real world is labor-intensive. As a result, the power of simulation is often harnessed to virtually generate 6D pose labels for training deep learning models \cite{self6d}. However, these methods typically exhibit inferior performance when applied to real world data, due to the persistent domain gap between real and synthetic data, as demonstrated in Fig. \ref{fig:fig1}. Many approaches \cite{Ssd-6d,PBR1} rely on a customized simulation strategy to produce high-quality synthetic data, while others \cite{gan1,feature_mapping} necessitate a specialized network architecture to extract domain-invariant signals. Alternatively, some methods employ render-and-compare techniques \cite{self6d,texpose} for self-supervision on unlabeled real data. These approaches utilize 2D appearance information of objects to distinguish reliable predictions. Nevertheless, texture-less objects provide essentially no 2D effective features, posing a challenge for these methods.

In this study, we introduce a \textbf{S}ymmetric-aware keypoint prediction and \textbf{D}omain adaptation \textbf{Net}work (SD-Net) for 6D object pose estimation in bin-picking scenarios. SD-Net is constructed based on point-wise keypoint regression and deep Hough voting. The inclusion of a voting mechanism equips our model with the capability to perform reliable keypoint detection in bin-picking scenarios. To slove the iusse of object symmetry, we propose a new keypoint selection and filtering algorithm when performing keypoint prediction. Subsequently, all equivalent keypoints are calculated according to equivalent rotation matrices. This selection method, which takes into account object symmetry class, significantly streamlines the network's task of location and bolsters the pose estimation performance. We also implement a keypoint filtering algorithm to choose the predicted keypoints with highest confidence before generating pose hypotheses.

In addressing domain gap, we propose a sim-to-real framework under a student-teacher learning scheme which can be generalized to texture-less objects. We initially train teacher model in a fully-supervised manner with abundant synthesized data. Subsequently, we use the teacher model to generate pseudo labels on real world data, and then these pseudo labels are used to update the student network. To facilitate robustness, we propose a tailored heuristic for 3D geometry pseudo labeling that relies on semi-chamfer distance, enabling the careful identification of reliable predictions. Moreover, the integration of mask labels contributes to the stability of the training process.

We benchmark our proposed method using the Sil\'eane \cite{s} and Parametric \cite{parametricnet} datasets. Experimental results demonstrate that SD-Net outperforms state-of-the-art methods. On the Sil\'eane dataset, SD-Net achieves a 6\% improvement in average precision. Meanwhile, on the Parametric dataset, SD-Net surpasses the state-of-the-art method by 8\% in terms of average precision. In summary, the main contributions of this work are:

\begin{itemize}
    \item We introduce a new 6D object pose estimation network with symmetric-aware keypoint prediction and domain adaptation, which achieves state-of-the-art estimation performance on the Sil\'eane and Parametric datasets. 
    \item We propose a new keypoint selection method that considers object symmetry class and a robust keypoint filtering algorithm that dynamically eliminates multiple and outlier keypoint candidates. 
    \item We propose an iterative self-training framework for domain adaptation in 6D object pose estimation, which leverages the 3D geometry information of objects to carefully distinguish pseudo labels.
\end{itemize}

\section{RELATED WORK} \label{introduction}
\subsection{Holistic Methods} 
Holistic methods directly estimate the pose of object and can be divided into two main groups (classical and learning-based methods). PPF \cite{PPF} proposes point pair global shape feature to retrieve poses from scene point cloud. Hinterstoisser \cite{Hinterstoisser1}, \cite{Hinterstoisser2} proposes a new feature which uses RGB color gradient and 3D surface normal information. These classical methods are not robust in clustered scenes. Based on deep neural network, some methods transform pose estimation problems into regression problems. DenseFusion \cite{densefusion} uses dense pixel level fusion to fuse RGB and point cloud. Some works \cite{p1,p2,MPP-Net,opnet1,opnet2,sock} directly regresses the position and rotation of the object from point cloud or depth map. Because the rotation space is nonlinear, direct regression of rotation is challenging.

\subsection{Keypoint-based methods}
Keypoint-based methods detect keypoints in camera coordinate system and establishe correspondence between them and keypoints in canonical object-frame coordinate system. In terms of keypoint selection, BB8 \cite{Bb8} and YOLO-6D \cite{YOLO-6D} predict the projection of the 3D bounding box on the 2D plane. The corners of the bounding box are generally far from the surface of the object and are less representative. Therefore, PVNet \cite{pvnet} and PVN3D \cite{pvn3d} use the farthest point sampling algorithm to select keypoint to reduce position errors. For symmetry objects, the points selected in this way are highly similar and difficult to distinguish. FFB6D \cite{ffb6d} uses SIFT \cite{SIFT} to extract features different angles 3D models and associate them with the corresponding 3D positions. However, these work \cite{ffb6d,ORB-FPS} is not suitable for texture-less objects. ParametricNet [4] predicts keypoints in shape templates which has two disadvantages. First, chirality problems \cite{parametricnet} occur when predicting keypoint of mirror symmetry objects. Second, the selection of keypoint according to object shape parameter is tedious and less robust. These methods fail to consider the geometric similarity of different components of symmetry objects, thus ignoring pose ambiguity caused by object symmetries. It is worth noting that there is a lack of a robust strategy to select keypoints from symmetry objects in keypoint prediction stage.

\begin{figure*}[t]
	\centering
		\includegraphics[width=1.9\columnwidth]{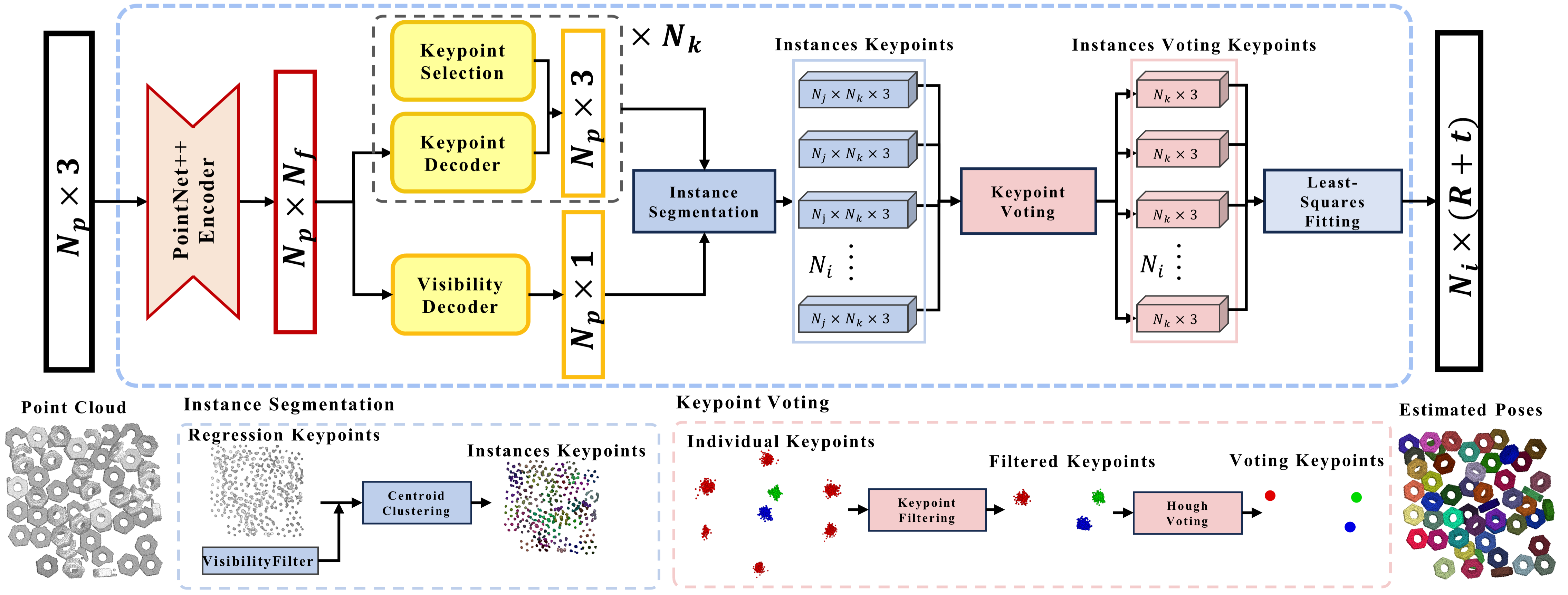}
	\caption{Overview of SD-Net architecture. SD-Net is constructed based on point-wise keypoint regression and deep Hough voting. It consists of two main parts: symmetri-caware keypoint prediction and self-supervised domain adaptation. keypoint prediction consists of a new keypoint selection and filtering algorithm. We omit the domain adaptation framework, for brevity and more details can be found in Section \ref{subsec: domain}. $N_j$ represents the number of point cloud points for each instance. $N_k$ represents the number of keypoint decoders and corresponds to the number of keypoint of objects. $N_i$ represents the number of instances in the scene.}
	\label{fig:fig2}
\end{figure*}
\subsection{Domain Adaptation for 6D Pose Estimation}
Sim-to-real transfer is crucial in 6D pose estimation as it bridges the domain gap between synthetic and real data observations through robotic visual systems. Some works from domain randomization aim at sampling a wide variety of simulation settings to learn domain-invariant attributes. such as random backgrounds \cite{Ssd-6d}, \cite{aae} and image augmentations \cite{image_augmentations}. Some works \cite{PBR1}, \cite{PBR2} harness light-weight physically-based renderer (PBR) data to simulate realistic texture to reduce the gap between the synthetic and real domains. Some works learning a mapping between different visual domains based on Generative Adversarial Networks \cite{gan1} or means of feature  \cite{feature_mapping} mapping. However, these methods either depend on a customized simulation strategy to generate high-quality synthetic data or require a specialized network architecture to extract domain-invariant signals. Alternatively, some methods \cite{self6d}, \cite{cps++++}, \cite{texpose} conduct render-and-compare strategy to self-supervised on unlabeled real data and impose consistencies between rendered features and sensed features. These methods harnesses 2D appearance information of objects to distinguish reliable predictions. However, texture-less objects provide essentially no 2D effective features, posing a challenge for these methods. In addition, these methods focus on pose estimation with domain adaptation of individual objects, ignoring the entire scene.

\section{Method}
Given a point cloud of a bin-picking scene where multiple object instances are stacked randomly into a pile, we are interested in detecting instances and estimating their rotation $R \in SO(3)$ and translation $t \in \mathbb{R}^{^{3}}$ in three-dimensional (3D) space. The object pose is represented by a rigid transformation from the object coordinate system to a reference camera coordinate system. In this section, we first introduce our symmetric-aware keypoint selection and filtering algorithm for SD-Net. Then, we present the overall architecture of SD-Net. In order to boost the performance of object pose estimation in real world, we introduce the self-supervised domain adaptation framework.

\subsection{Symmetric-Aware keypoint prediction}
\begin{figure}[t]
	\centering
		\includegraphics[width=1.0\columnwidth]{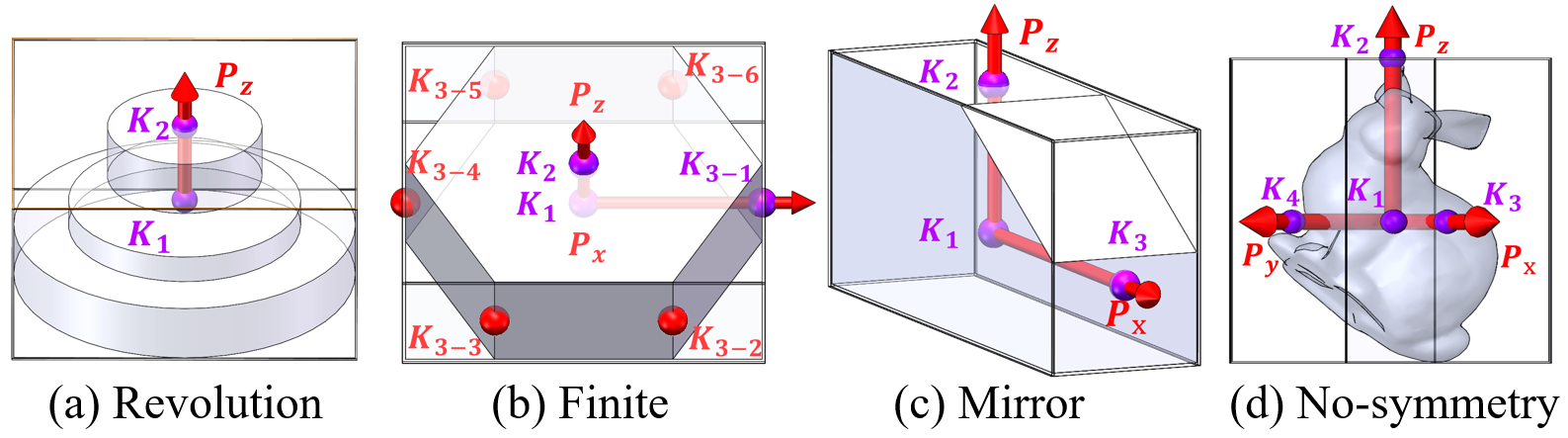}
	\caption{The axes selection strategy depends on the object symmetry class. The red axes represent the selected axes. The blue dots represent the selected keypoints, while the red dots represent the corresponding equivalent keypoints.}
	\label{fig:fig3}
\end{figure}

\textbf{Keypoint Selection}. 
The selection of keypoint is a significant challenge. Keypoints away from the surface of the object will increase localization errors \cite{pvnet}. It is difficult to distinguish points on a symmetry object, which means that the geometric features of different components of the object are similar. To facilitate network convergence, the keypoint selection strategy should meet the following requirements: (1) Keypoints should be close to the surface of the object. (2) Take the object symmetry class into account. Based on this observation, we propose a heuristic keypoint selection algorithms. Keypoint set is a collection of a object centroid and the points where the object bounding box intersect with the object coordinate axes. In this way, these selected key points are close to the surface of the object, making point-based networks easy to aggregate scene context in the vicinity of them. Keypoint set is defined as:
\begin{align}
	K = {P_c} \cup \{ {P_i}|{P_i} = {P_{bounding{\rm{ }}box}} \cap {P_{axles}}\} 
\end{align}
where $P_{c}$ represents the centroid, $P_{bounding box}$ represents the object bounding box and $P_{axes}$ represents the object selected coordinate axes. For example, in Fig. \ref{fig:fig3}.a, the keypoint set contains elements $\{ {K_1},{K_2}\} $. Subsequently, an adaptive coordinate axes selection strategy considering objects symmetry class designed and the details are as follows:
\label{subsec: chirality}  
\begin{itemize}
	\item For revolution symmetry objects, we choose the axis of rotation as ${P_{axes}}$. In this way, it can avoid selecting points located on the curved surface which have a low discrimination.
	\item For finite non trivial symmetry objects, we choose the axis of rotation and another axis which is randomly selected from the remaining axes as ${P_{axes}}$.
	\item For mirror symmetry objects, we choose two axes which parallel to the plane of mirror symmetry as ${P_{axes}}$. This can avoid chirality issues, which is that two point sets of different chirality structures cannot be registered by rotation and translation transformation. 
 	\item For no proper symmetry objects, we select all three axes as ${P_{axes}}$. This can increase the number of keypoints and improve pose prediction accuracy and robustness.
  
\end{itemize}

In addition, for revolution and finite symmetric objects, we transform the object model so that any coordinate axis of the objects coincides with the rotation axis. For mirror symmetry objects, we transform the object model so that any two coordinate axes of the objects coincide with the symmetry plane. Any object instances can be divided into any of the four symmetry class above it and this keypoint selection can be used for any type of objects.

\textbf{Equivalent Keypoint Set}. In the process of selecting keypoint using Equation (1), some keypoints are often accompanied by multiple equivalent keypoints, especially for finite symmetric object. For example, in Fig. \ref{fig:fig3}.b, the keypoint ${K_{3}}$ formed by the intersection of the ${X}$-axis and the bounding box contains six equivalent keypoints $ \{ {K_{3 - 1}},{K_{3 - 3}},...,{K_{3 - 6}}\}$. These keypoints remain geometrically consistent, which can cause ambiguity during the training phase of the network. Therefore, we predict all equivalent keypoints to avoid confusion during the learning phase. Equivalent keypoint set is defined as:
\begin{align}
	{K_{j - equivalent}} = \{ {K_j}g|g \in G\} 
\end{align}
where $G \in SO(3)$ is the set of rotation matrix that keeps object state unchanged. For no proper symmetry objects, $G$ is the unit matrix. ${K_j}$ is an element in set ${K}$ in Equation (1).

\begin{figure}[h]	
	\centering
	\subfigure[Keypoint selection]{	
		\begin{minipage}[b]{0.30\columnwidth}
			\includegraphics[width=1\columnwidth, height=2.2cm]{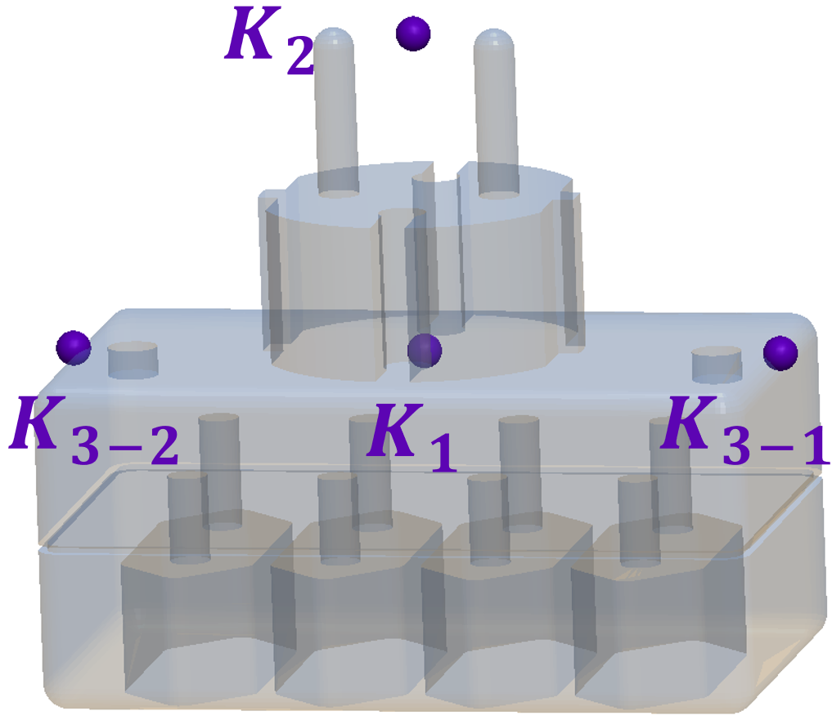}
		\end{minipage}
		\label{fig:fig4a}
		}
	\subfigure[Keypoint regression]{\centering
		\begin{minipage}[b]{0.30\columnwidth}
			\includegraphics[width=1\columnwidth, height=2.2cm]{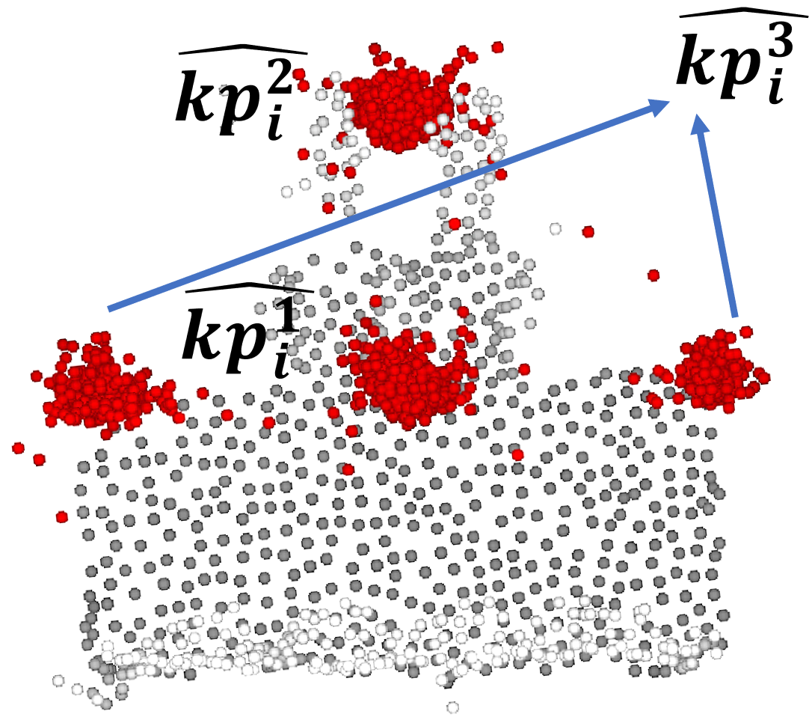}
		\end{minipage}
		\label{fig:fig4b}
		}		 
	\subfigure[Keypoint filtering]{\centering
		\begin{minipage}[b]{0.30\columnwidth}
			\includegraphics[width=1\columnwidth, height=2.2cm]{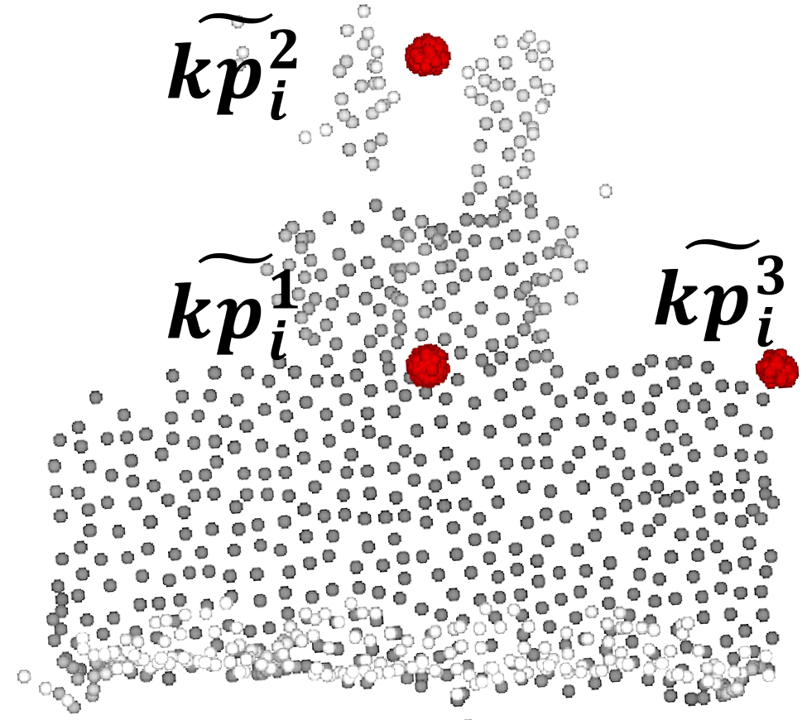}
		\end{minipage}
		\label{fig:fig4c}
		}	
	\caption{(a) Keypoints sampled by ours selection algorithms on t-less20 object from Sil\'eane dataset. (b) The white points represent the scene instance point clouds, and the red points represent the pointwise predicted keypoints. (c) The red points represent predicted keypoints after filtering.}
	\label{fig:fig4}
\end{figure}

\textbf{Keypoint Filtering} In the model inference stage (as described in Section \ref{subsec: Pose Hypotheses}), some predictied keypoints will be distributed into multiple candidate clusters around equivalent keypoints lable, so as the predicted ${\widehat {kp_i^3}}$ in Fig. \ref{fig:fig4}.b. Moreover, some prediction keypoints have significant deviation from ground truth. 

Given the predicted keypoint, we are ready to eliminate multiple ambiguity and outlier keypoint candidates. We introduce a robust keypoint filtering algorithm. It is detailed in Algorithm \ref{algorithm1}. The core step of our algorithm is to cluster the predicted keypoints to form multiple keypoint cluster and find the cluster with the highest point cloud density. To achieve $cluster$ in Algorithm \ref{algorithm1}, we employ DBSCAN \cite{DBSCAN}. The $density$ in Algorithm \ref{algorithm1} is used to estimate the density of point cloud clusters and is defined as:
\begin{align}
D = \frac{1}{N}\sum\limits_{i = 1}^N {{{\left\| {{p_i} - {p_c}} \right\|}_2}} 
\end{align}
where ${p_i}$ and ${p_c}$ represent the points and centroids in the point cloud clusters, respectively. 

As shown in Fig. \ref{fig:fig4}.c, keypoint filtering algorithm provides reliable keypoint to fit the pose. For predictied keypoint ${\widehat {kp_i^3}}$ distributed into two candidate clusters, keypoint filtering algorithm dynamically preserves the candidates clusters in the area with the highest density. Meanwhile, the outliers keypoint with significant prediction deviations can be considered as low density point clouds, so they can be filtered out by the density threshold. It is worth noting that we apply keypoint filtering to each type of keypoint respectively, because different types of keypoint are distributed differently.

\renewcommand{\algorithmicrequire}{\textbf{Input:}}  
\renewcommand{\algorithmicensure}{\textbf{Output:}} 

\begin{algorithm}
    \caption{Density-based Keypoint Filtering}
    \begin{algorithmic}[1]
        \Require prediction keypoints ${P_1} = \{ {x_1},{x_2},...,{x_n}\}  \in {^{n \times 3}}$.
        \Ensure filtered keypoints ${P_2} = \{ {x_1},{x_2},...,{x_m}\}  \in {^{m \times 3}}$.
        
        \State initialize an empty cluster set $C$
        \State  $C = cluster({P_1})$
        \State  ${d_0} = density({C_0})$
        \State  ${P_2} = C_0$

        \For{${C_i}$ in ${C}$}
              \If {$density({C_i}) < {d_0}$} 
                \State ${P_2} \leftarrow {C_i}$
                \State ${d_0} \leftarrow density({C_i})$
             \EndIf
         \EndFor
        \State \Return{${P_2}$}

    \end{algorithmic}
    \label{algorithm1}
\end{algorithm}

\subsection{Architecture design}
\label{subsec: net}
Fig. \ref{fig:fig2} illustrates our end-to-end 6D Pose Estimation network. Initially, it takes the point clouds of the bin-picking scene with $N_p$ points as input and applies a feedforward network PointNet++ \cite{pointnet++} for feature extraction. The extracted $F_{p}$ has a size of $N_p\times N_f$. Subsequently, our network diverges into two decoders which consume $F_{p}$ to predict the keypoint and visibility for each individual point. 

\textbf{Visibility Decoder}. In bin-picking scenarios, some instances are severely occluded. We are not interested in these instances which cannot be captured at the bottom. Therefore, we set the visibility decoder to reduce the impact of severe occlusion. The point-wise visibility is defined as ${V_i} = {N_i}/{N_{\max }}$. ${N_i}$ is the number of points of the instance to which the $i$ th point belongs and ${N_{\max }}$ indicate the highest number of points within all visiable instances. We pass $F_p$ into visibility decoder to prediction the point-wise visibility. The visibility loss is defined as:
\begin{align}
{L_v} = \frac{1}{{{N_p}}}\sum\limits_{i = 1}^{{N_p}} {\left\| {{{\widehat V}_i} - {V_i}} \right\|} 
\end{align}
where ${\widehat V_i}$ denotes the prediction of the point-wise visibility.

\textbf{Keypoint Decoder}. Compared with directly predicting keypoint, predicting the offset of keypoint relative to the point cloud is more accurate. Therefore, we can pass $F_p$ into a keypoint decoder to predict the point-wise offsets ${o_i} \in \mathbb{R}^{^{3}}$. The predicted point-wise keypoint coordinates can be expressed as $\widehat {{kp_i}} = {o_i} + {p_i}$, where ${p_i}$ denotes the $i$ th point coordinate. Different types of keypoints are predicted separately using independent decoders. For objects with ${N_k}$ types of keypoints, we use ${N_k}$ decoders to obtain prediction keypoints $\widehat {kp_i^j} \in \mathbb{R}^{^{3}}$, which denotes the prediction of the $j$ th keypoint of the instance to which the $i$ th point corresponds. The keypoint prediction loss is defined as:
\begin{align}
	{L_k} = \frac{1}{{{N_p}}}\sum\limits_{j = 1}^{{N_k}} {\sum\limits_{i = 1}^{{N_p}} {\mathop {\min }\limits_{kp \in kp_i^j} \left\| {\widehat {kp_i^j} - kp} \right\|} }
\end{align}
where $kp_i^j$ denotes the point-wise keypoint labels ($kp_i^j \in \mathbb{R}^{^{{N_e} \times 3}}$, $N_e$ denotes the number of equivalent keypoint). This loss function computes the Euclidean distance between the prediction and the closest ground truth. 

\textbf{Generating Pose Hypotheses}. \label{subsec: Pose Hypotheses} In the inference stage, as shown in Fig. \ref{fig:fig2}, we filter out severely occluded scene point clouds and group scene points into instances with Mean Shift algorithm \cite{meanshift} to generate instances keypoints. Then we apply keypoints filtering and hough voting to generate each instance voting keypoints $\{ \widehat {{K_j}} \in \mathbb{R}^{^{3}}\} _{j = 1}^{{N_k}}$ in the camera coordinate system. When providing their corresponding points $\{ {K_j}\} _{j = 1}^{{N_k}}$ in canonical object coordinate system, we utilize a least-squares fitting algorithm \cite{least} to calculate the optimal values of ${R}$ and ${t}$.
 \begin{align}
 {L_{lsf}} = \sum\limits_{j = 1}^{{N_k}} {{{\left\| {\widehat {{K_j}} - (R \cdot {K_j} + t)} \right\|}_2}}  
\end{align}

\begin{figure}[t]
	\centering
		\includegraphics[width=1.0\columnwidth]{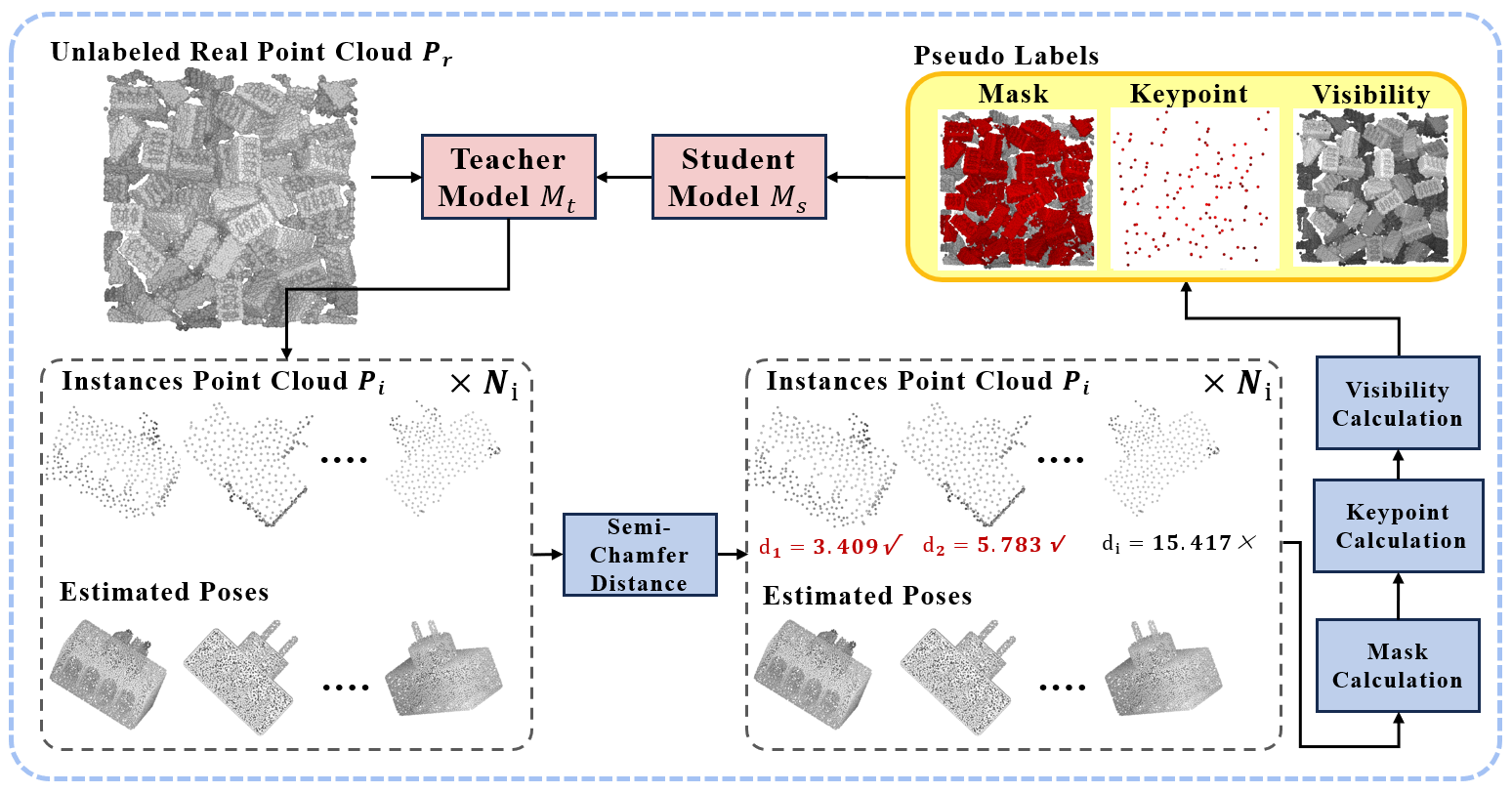}
	\caption{Overview of our proposed self-supervised domain adaptation framework for 6D object pose estimation. We first train teacher model on synthesize abundant data to generate initial pose predictions. We then use a 3D geometry pseudo labelling algorithm to distinguish real word predictions for student model training. In the next iteration, the teacher model is initialized as the last trained student model and iterate the above process util the model convergence.}
	\label{fig:fig8}
\end{figure}

\subsection{Self-supervised domain adaptation}
\label{subsec: domain}
Fig. 5 summarizes our student-teacher domain adaptation pose estimation framework. We first train a fully-supervised teacher model $M_{t}$ with abundant labeled synthetic data, as introduced in Section \ref{subsec: net}. Then we apply $M_{t}$ on unlabeled real data to predict object instances poses. Based on these initial poses with decent quality, a robust label selection algorithm is designed to select the best reliable predictions for pseudo labels. The real data with pseudo label generated by pose prediction is utilized to train a student model $M_{s}$ by self-supervised learning. Crucially, we iterate the above process by taking the $M_{s}$ as a new teacher model, in order to progressively boost the quality of pseudo labels and close the domain gap.

Specifically, given unlabeled real point cloud $P_r$ and their initial pose estimations $\{ {p_i} = [{R_i}|{t_i}]\} _{i = 1}^{N_i}$ generated by teacher model, we harness geometric constraints to seek the best alignment w.r.t. 6D pose. The core idea is to generate the object point cloud that corresponds to the predicted pose and compare with the real collected point cloud to determine whether the predicted pose is reliable or not. We conduct the following two steps to leverage geometric constraints. We first backproject the object CAD model $C$ using the corresponding predicted pose to retrieve the point clouds $C_i$ in camera space: ${C_i} = {R_i} \times C + {t_i}$. Object instances point clouds $P_i$ only contains the surface point cloud that is visible from a particular viewpoint and severely obstructed. $P_i$ is incomplete point cloud and a subset of $C_i$. Intuitively, we use the semi-chamfer distance between $P_i$ and $C_i$ as a 3D metric to quantify the quality of predicted pose:
\begin{align}
	{d_i} = \frac{1}{{{N_{{P_i}}}}}\sum\limits_{x \in {P_i}}^{} {\mathop {\min }\limits_{y \in {C_i}} \left\| {x - y} \right\|{}_2} 
\end{align}
where $x$ and $y$ denotes 3D points from $P_i$ and $C_i$ respectively, ${N_{{P_i}}}$ denotes the number of points for $P_i$.

We calculate the pose quality for each predicted poses generated by tearcher model $M_{t}$ and obtain geometry pose quality set $\{ {d_i}\} _{i = 1}^{{N_i}}$ in a bin-picking scene. Then, we dynamically generate a threshold $d_g$ based on the mean and standard deviation of the geometry pose quality set distribution. Pose prediction $p_{i}$ are regarded as the correct prediction when ${d_i} < {d_g}$. The pseudo labels of keypoints for the objects in the real data are calculated according to the correct pose predictions. The pesudo labels of visibility are calculated based on the number of each object instance point cloud. Additionally, for instance point clouds whose geometry pose quality ${d_i} > {d_g}$, the pesudo labels of mask are assigned to exclude them from the model training. Specifically, only pesudo labels with a mask label of 1 participate in training. After label generation, we incorporate real point clouds with reliable pseudo labels and train a student model $M_{s}$ to transfer the knowledge from the synthetic data to real data. In the next iteration, the teacher model $M_{t}$ is initialized as the last trained student model $M_{s}$ and iterate the above process util the model convergence.

\section{Experiments} 
\begin{figure*}[t]
	\centering
		\includegraphics[width=2\columnwidth]{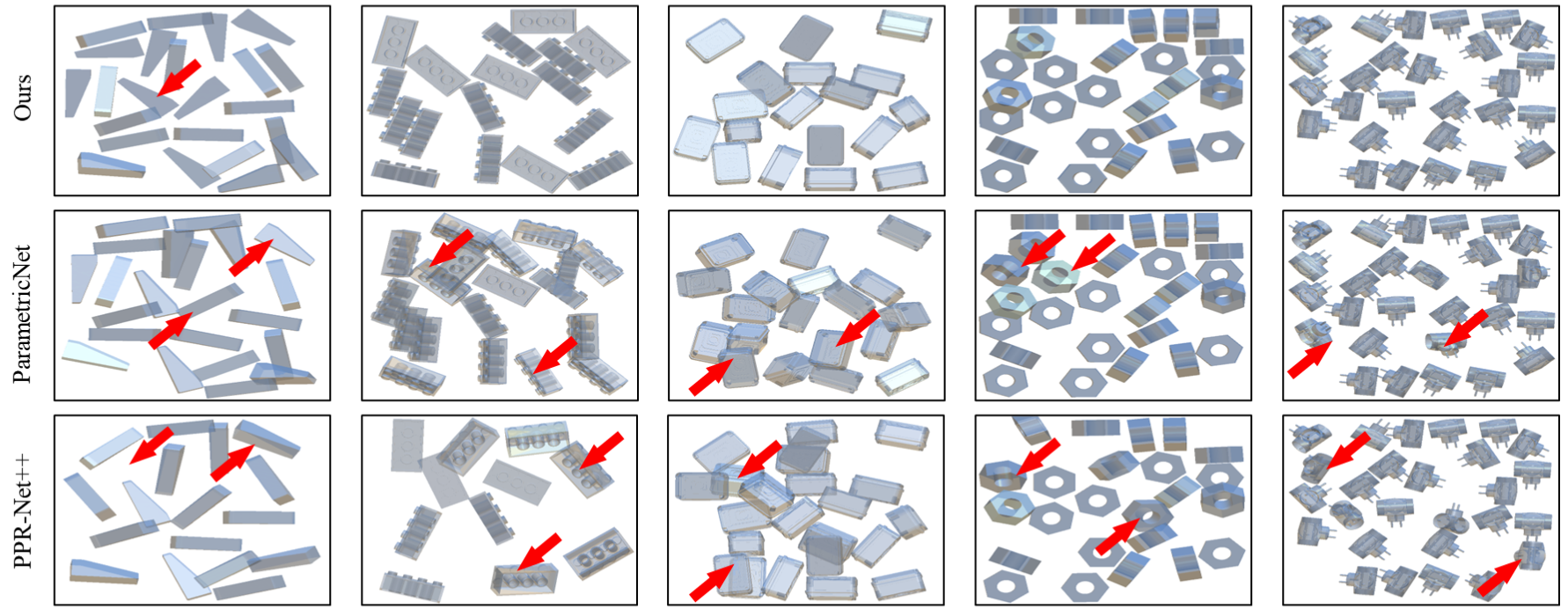}
	\caption{Qualitative results on Sil\'eane dataset \cite{s}. Rows show a comparison against different methods (SD-Net, ParametricNet and PPR-Net++). Columns show different scenarios. Red arrows highlight wrong prediction pose and we highlight a maximum of 2 wrong pose in a scene for brevity.}
	\label{fig:fig5}
\end{figure*}

\begin{table*}[htb]
	\centering
		\caption{ Quantitative evaluation of 6D Pose estimation on Sil\'eane dataset \cite{s}. The test objects contain three types of symmetry.}
		\begin{tabular}{c|c c c c c c c c|c}
		\Xhline{0.5pt}
~Object                   & Bunny            & C.Stick       & Pepper        & Brick         & Gear          & T-Less20      & T-Less22      & T-Less29  & Mean     \\ 
~Symmetry Class   & Non-Symmetry & Revolution & Revolution & Finite & Revolution & Finite & Non-Symmetry & Finite  &      \\ \hline
PPF \cite{PPF}           & 0.29          & 0.16          & 0.06          & 0.08          & 0.62          & 0.20          & 0.08          & 0.19      & 0.21    \\
LINEMOD+ PP\cite{Hinterstoisser2}      & 0.45          & 0.49          & 0.03          & 0.39          & 0.50          & 0.31          & 0.21          & 0.26    & 0.33      \\
Sock et al \cite{sock}    & 0.74          & 0.64          & 0.43          & -             & -             & -             & -             & -          & -   \\
PPR-Net with ICP \cite{p1}       & 0.89          & 0.95          & 0.84          & -             & -             & 0.85          & -             & -            & - \\
OP-Net with Lori$_1$ \cite{opnet1}       & 0.92          & 0.94          & 0.98          & 0.41          & 0.82          & 0.85          & 0.77          & 0.51         & 0.78  \\
OP-Net AP \cite{opnet2}       & 0.92          & 0.98          &\textbf{0.99}          & 0.45          & 0.82          & 0.87          & 0.84          & 0.56     & 0.80     \\
ParametricNet \cite{parametricnet} & -             & 0.97          & -             & -             & \textbf{1.00} & 0.92          & -             & 0.94   & -       \\
PPR-Net++ \cite{p2}    & 0.99          & 0.98          & 0.98          & 0.47          & \textbf{1.00} & 0.93          & 0.92          & 0.94        & 0.90 \\
Ours          & \textbf{1.00} & \textbf{1.00} & \textbf{1.00} & \textbf{0.75} & \textbf{1.00} & \textbf{0.98} & \textbf{0.96} & \textbf{0.98}   & \textbf{0.96}\\

		\Xhline{0.5pt}
		\end{tabular}
	\label{tab:table1}
\end{table*}

\subsection{Datasets and evaluation metrics}
To comprehensively evaluate our method in bin-picking scenarios, we select Sil\'eane dataset \cite{s} and Parametric dataset \cite{parametricnet}. Sil\'eane dataset is comprised of a total of more than 2,600 bin-picking scenarios. Parametric dataset consists two types of data. The L-dataset test set consists of objects with the same parameters as the training set, used to evaluate learning abilities. The G-dataset test set consists of objects with different parameters compared to the training set, used to assess generalization abilities.

This metric evaluates the performance of methods by calculating the area under the precision-recall curve, which is summarized as the Average Precision. Specifically, when the distance between predicted pose and ground truth is less than 0.1 times object's minimum bounding sphere diameter, the prediction pose is considered correct \cite{bregier}. 

\subsection{Evaluation on Sil\'eane dataset}

Table \ref{tab:table1} summarizes the comparison results between our approach and current 6D pose estimation methods. Our proposed approach achieves state-of-the-art results and outperforms others, obtaining an average precision of 96\%. Our approach results in an improvement of +68\% to +75\% on average precision compared with the conventional approaches \cite{PPF,Hinterstoisser2}. Our approaches already clearly outperforms the existing deep learning-based methods \cite{opnet1,opnet2,parametricnet,p2} with a large margin without the need of a separate refinement stage. Fig. \ref{fig:fig5} shows the qualitative results of SD-Net in severely bin-picking scenarios and SD-Net is superior to the other two methods in handling symmetry objects.

The observed improvement in finite non trivial symmetry objects is likely due the equivalent keypoint set which is easier to learn by the regression network, e.g., T-Less20 and T-Less29 objects. For the brick object, compared with the current state-of-the-art PPR-Net++, SD-Net has a significant increase of 23\% in average precision. We observe that smaller object sizes, result in larger predicted position and rotation deviations. Our keypoint filtering algorithm eliminate the keypoints with significant prediction deviation, which helps to improve the accuracy of pose estimation.

\subsection{Evaluation on Parametric dataset}

\textbf{Learning ability}. From Table \ref{tab:table2}, SD-Net advances state-of-the-art results by 8\% on average precision metric. Compared with ParametricNet on TN42 object, SD-Net significantly improves the performance by 26\%. A crucial contributing factor is that our keypoints selection algorithm avoid chirality issues (as described in Section \ref{subsec: chirality}). TN06 object has 12 equivalent pose, resulting in giving rise to ambiguous pose estimations. SD-Net conduct equivalent keypoint set and density based keypoint filtering algorithm to reduce the pose estimation ambiguity and further improve the pose accuracy. In TN16 and TN34 objects scenario, the level of occlusion is relatively low, resulting in highly accurate prediction outcomes for all three methods.

\begin{table}[htb]
	\centering
		\caption{Learning ability evaluation on Parametric dataset.}
		\begin{tabular}{c|c c c c|c}
		\Xhline{0.5pt}
		Object  	& TN06	&	TN16	&	TN34	&	TN42	&	Mean	\\ 
		Symmetry Class	& Finite	&	Revolution	&	Revolution	&	Mirror	&	 	\\ \hline
  
		PPR-Net++ 	&	0.80	&	0.99	&	\textbf{1.00}	&	0.39	&	0.80	\\
      	ParametricNet	&	0.94	&	\textbf{1.00}	&	\textbf{1.00}	&	0.52	&	0.87	\\
		Ours	& \textbf{1.00}	&	\textbf{1.00}	&	\textbf{1.00}	&	\textbf{0.78}  &	\textbf{0.95}	\\
		\Xhline{0.5pt}
		\end{tabular}
	\label{tab:table2}
\end{table}

\textbf{Generalization ability}. 
In Table \ref{tab:table3}, our average precision metric is 8\% higher than the state-of-the-art method. SD-Net demonstrates almost the same generalization capability as its learning capability. The results show that SD-Net exhibits strong generalization capability for unseen parameters objects. To further evaluate the SD-Net generalization ability, we conduct experiments by down-sampling the number of training instances to one-third and one-fifth of the original dataset, which means that the model has seen fewer parameters objects during training. Across varying initialization experimental configuration, SD-Net consistently outperforms other methods on generalization metric. For the TN06 and TN42 object, SD-Net achieves even higher average precision when trained with only one-fifth of the objects compared to other methods trained on all objects.

\begin{table}
	\centering
		\caption{Generalization ability evaluation on Parametric dataset.}
		\begin{tabular}{c|c| c c c c  c}
		
		\Xhline{0.5pt}
		
		\multirowcell{2}{Train \\ Mode}  & \multirowcell{2}{Method} & \multirowcell{2}{TN06} &\multirowcell{2}{TN16} & \multirowcell{2}{TN34} & \multirowcell{2}{TN42}  & \multirowcell{2}{Mean}\\ &\\ 
		\hline

		\multirowcell{3}{Learn \\ all}
				& PPR-Net++			& 0.79	& 0.99	& \textbf{1.00}	& 0.28	& 0.77\\
				& ParametricNet		& 0.93	& \textbf{1.00}	& \textbf{1.00}	& 0.51	& 0.86\\
				& Ours	& \textbf{1.00}	& \textbf{1.00}	& \textbf{1.00}	& \textbf{0.77}	& \textbf{0.94}\\
		\hline
		
		\multirowcell{3}{Learn \\ 1/3}
				& PPR-Net++				& 0.78	& 0.94	& 0.96	& 0.22	& 0.73\\
				& ParametricNet		& 0.86	& 0.98	& \textbf{1.00}	& 0.41	& 0.81\\
				& Ours	& \textbf{1.00}	& \textbf{1.00}	& \textbf{1.00}	& \textbf{0.67}	& \textbf{0.92}\\
		\hline
		
		\multirowcell{3}{Learn \\ 1/5}
				& PPR-Net++				& 0.77	& 0.56	& 0.83	& 0.18	& 0.59\\
				& ParametricNet		& 0.86	& 0.63	& 0.86	& 0.39	& 0.69\\
				& Ours	& \textbf{0.98}	& \textbf{0.65}	& \textbf{0.89}	& \textbf{0.54}	& \textbf{0.77}\\
		\hline

		\Xhline{0.5pt}
		\end{tabular}
	\label{tab:table3}
\end{table}

\begin{table}[htb]
	\centering
		\caption{Ablation study on Sil\'eane dataset.EKS, Equivalent Keypoint Set. KF, Keypoint Filtering. SCD, Semi-Chamfer Distance. DA, Domain Adaptation. w/o,without.}
		\begin{tabular}{c|c|c|c|c}
		\Xhline{0.5pt}
			Object	&	Brick	&	T-Less20	&	T-Less22	&	T-Less29	\\	\hline
			SD-Net (w/o EKS)	& 0.08	&	0.76	&	\textbf{0.95}	&	0.66	\\
			SD-Net (w/o KF)		& 0.53	&	0.03	&	0.94	&	0.08	\\
                SD-Net (w/o DA) 	& 0.70	&	0.96	&	0.95	&	0.97	\\	
                SD-Net (w/o SCD)    & 0.71	&	0.93	&	0.94	&	0.00	\\
                SD-Net (BBox)		& 0.21	&	0.55	&	0.61	&	0.16	\\
   			SD-Net (FPS)		& 0.45	&	0.89	&	0.88	&	0.79	\\
   			SD-Net (Parameter)		& 0.47	&	0.92	&	0.92	&	0.94	\\
			SD-Net	&	\textbf{0.75}	&	\textbf{0.98}	&	\textbf{0.96}	&	\textbf{0.98}	\\
		\Xhline{0.5pt}
		\end{tabular}
	\label{tab:table4}
\end{table}

\subsection{Ablation study}
We present extensive ablation studies on SD-Net to compare different design choices. Table \ref{tab:table4} summaries the evaluation results on the Sil\'eane dataset. SD-Net (w/o EKS) learns the keypoint without equivalent keypoint set. For symmetric objects, its performance is greatly reduced. SD-Net (w/o KF) remove keypoint filtering algorithm and the performance degradation is large. SD-Net (w/o DA) remove domain adaptation and the results show that ours self-training domain adaptation framework enhance the SD-Net performance on real data. SD-Net (w/o SCD) use chamfer distance to quantify the quality of prediction pose. Due to the incomplete object instances point clouds, chamfer distance cannot precisely quantify the quality. We replace the symmetric-aware keypoint selection strategy with BBox, FPS, and Parameter and does not perform a keypoint filtering algorithm. The performance of these methods is greatly reduced. This shows that our proposed symmetric-aware keypoint prediction performs reliable detection keypoints. In general, the proposed novelty design can significantly improve average precision.

\subsection{Real World Experiment}
We further explore the application of SD-Net to robot grasping tasks in the real world. We choose TN06 object instance from Parametric \cite{parametricnet} datasets as the grab object. We utilize the Blender platform to generate simulation data set. Totally 18000 point clouds are annotated, comprising 300 cycles, with each cycle consisting of 60 scenarios. In addition, we collected 600 unlabeled real world point clouds to self-training domain adaptation. 
\begin{figure}[h]	
	\centering
	\subfigure[Scene point clouds]{	
		\begin{minipage}[b]{0.30\columnwidth}
			\includegraphics[width=1\columnwidth, height=2cm]{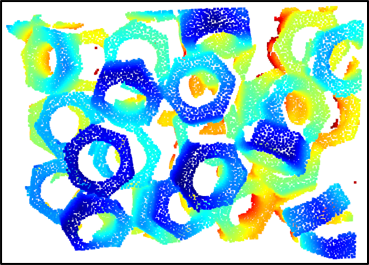}
		\end{minipage}
		\label{fig:fig7a}
		}
	\subfigure[Predict object pose]{\centering
		\begin{minipage}[b]{0.30\columnwidth}
			\includegraphics[width=1\columnwidth, height=2cm]{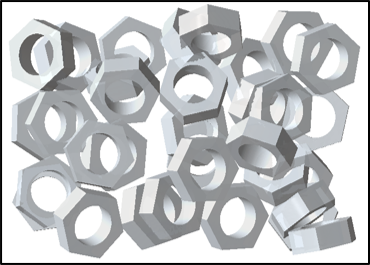}
		\end{minipage}
		\label{fig:fig7b}
		}		
	\subfigure[Grasp pose]{\centering
		\begin{minipage}[b]{0.30\columnwidth}
			\includegraphics[width=1\columnwidth, height=2cm]{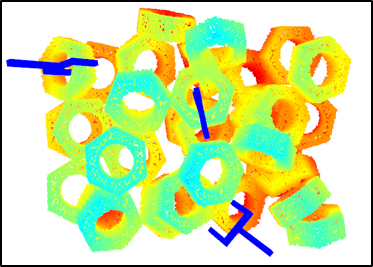}
		\end{minipage}
		\label{fig:fig7c}
		}

	\subfigure[Scene RGB image]{	
		\begin{minipage}[b]{0.30\columnwidth}
			\includegraphics[width=1\columnwidth, height=2.0cm]{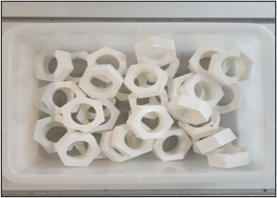}
		\end{minipage}
		\label{fig:fig7d}
		}
	\subfigure[Experiment setup]{\centering
		\begin{minipage}[b]{0.30\columnwidth}
			\includegraphics[width=1\columnwidth, height=2.0cm]{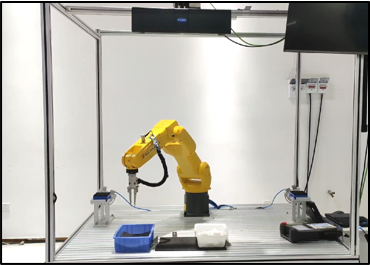}
		\end{minipage}
		\label{fig:fig7be}
		}		
	\subfigure[Robotic grasp]{\centering
		\begin{minipage}[b]{0.30\columnwidth}
			\includegraphics[width=1\columnwidth, height=2.0cm]{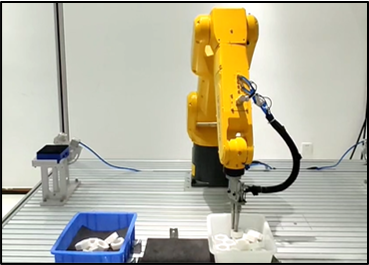}
		\end{minipage}
		\label{fig:fig7f}
		}

	\caption{Robotic grasping experiment on real world  bin-picking scenarios with the proposed approach SD-Net.}
	\label{fig:fig7}
\end{figure}

We deploy the aforementioned well-trained SD-Net on the Fanuc industrial robot, which is equipped with pneumatic gripper. The whole robot grasping system is implemented using the ROS and MoveIt! frameworks. In the real grasping experiment, objects are stacked in the container, As shown in Fig. \ref{fig:fig7} (d). The scene point clouds captured by the RVC X 3D camera is cropped, sampled, filtered, and subsequently fed into SD-Net for pose estimation, as shown in Fig. \ref{fig:fig7} (b). Based on the predicted instance pose, we select the set of grasp configurations from the pre-calculated grasps database, as shown in Fig. \ref{fig:fig7} (c). We evaluated the ability of SD-Net in 10 grasping trials. Our pipeline can successfully accomplish the robot grasping tasks for all graspable object instances in all trials. It show that SD-Net demonstrates excellent performance in robot bin-picking tasks.

\section{Conclusions}
In this paper, we propose a new 6D pose estimation network with symmetri-caware keypoint prediction and domain adaptation. It includes two critical components. We propose a new selection keypoint method which considers objects symmetry class and a robust keypoint filtering algorithm. We propose an network-agnostic iterative self-training framework for domain adaptation 6D object pose estimation. Experiments show that SD-Net has significant improvements in average precision compared to state-of-the-art approaches on the public Sil\'eane dataset and Parametric dataset.


\bibliographystyle{IEEEtran}
\bibliography{IEEEabrv,ParametricNet}









\end{document}